# YOLO9tr: A Lightweight Model for Pavement Damage Detection Utilizing a Generalized Efficient Layer Aggregation Network and Attention Mechanism

Sompote Youwai[1]
Achitaphon Chaiyaphat[1]
Pawarotorn Chaipetch[2]

[1] AI Research Group, Department of Civil Engineering, King Mongkut's University of Technology Thonburi
[2] Infraplus Co., Ltd., Bangkok
Email: Sompote.you@kmutt.ac.th

**Abstract:**
Maintaining road pavement integrity is crucial for ensuring safe and efficient transportation. Conventional methods for assessing pavement condition are often laborious and susceptible to human error. This paper proposes YOLO9tr, a novel lightweight object detection model for pavement damage detection, leveraging the advancements of deep learning. YOLO9tr is based on the YOLOv9 architecture, incorporating a partial attention block that enhances feature extraction and attention mechanisms, leading to improved detection performance in complex scenarios. The model is trained on a comprehensive dataset comprising road damage images from multiple countries, including an expanded set of damage categories beyond the standard four. This broadened classification range allows for a more accurate and realistic assessment of pavement conditions. Comparative analysis demonstrates YOLO9tr's superior precision and inference speed compared to state-of-the-art models like YOLO8, YOLO9 and YOLO10, achieving a balance between computational efficiency and detection accuracy. The model achieves a high frame rate of up to 136 FPS, making it suitable for real-time applications such as video surveillance and automated inspection systems. The research presents an ablation study to analyze the impact of architectural modifications and hyperparameter variations on model performance, further validating the effectiveness of the partial attention block. The results highlight YOLO9tr's potential for practical deployment in real-time pavement condition monitoring, contributing to the development of robust and efficient solutions for maintaining safe and functional road infrastructure.

## 1. Introduction

Within the sphere of transportation infrastructure, the structural integrity of road pavements is critical for ensuring commuter safety and enabling seamless operational flow. Substandard pavement conditions lead to a reduction in road capacity, characterized by decreased vehicle speeds and an increased risk of accidents due to compromised pavement integrity. Consequently, agencies tasked with pavement maintenance require effective methods for evaluating pavement conditions. Traditional approaches to assessing pavement deterioration are fraught with challenges, including labor-intensive processes and susceptibility to errors in human judgment. The assessment of road pavement conditions culminates in the calculation of the Pavement Condition Index (PCI) [1]. Developing a systematic approach for the accurate detection and categorization of pavement defects would enable engineers to perform timely and efficient pavement condition assessments. Utilizing such assessments, strategic interventions can be planned to improve road conditions, thus restoring them to an acceptable standard. To overcome these challenges, the present study proposes a novel approach that harnesses advancements in deep learning technologies for enhanced pavement condition analysis.

With the advent of advanced computational technologies, image processing methodologies have increasingly become integral to object detection within digital images. The emergence of artificial intelligence has catalyzed the creation of sophisticated automated detection and classification systems. These AI-driven target detection frameworks are fundamentally categorized into one-stage and two-stage detection algorithms. Two-stage algorithms, exemplified by R-CNN [2] Mask-R-CNN [3] and its subsequent iterations, commence with proposal generation prior to target detection. This methodology provides a measured response to class imbalance dilemmas, albeit at the cost of reduced speed, despite aspirations for heightened accuracy. Conversely, one-stage algorithms, such as the YOLO (You Only Look Once) series, directly ascertain the location and classification of targets within the image, thereby achieving remarkable detection velocities. Within the realm of real-time object detection, the YOLO algorithmic suite has undergone a significant evolution, characterized by advancements in detection rapidity, accuracy, and computational thrift. The pioneering YOLOv1 [4] algorithm instigated a transformative wave in the field through the deployment of a unified convolutional neural network that concurrently predicts bounding boxes and class probabilities. Ensuing versions, YOLOv2 [5]and YOLOv3 [6], introduced anchor boxes and multi-scale detection capabilities, respectively, markedly enhancing the detection efficacy for objects of diverse dimensions. Further iterations, YOLOv4 [7] and YOLOv5[8], honed the architectural framework, optimizing computational dispatch without detracting from precision. Progress persisted with YOLOv6[9] through YOLOv9[6, 10], which amalgamated cutting-edge methodologies from the expansive machine learning domain to augment model performance. The most recent iteration, YOLOv10 [11], has instigated a paradigmatic shift by eliminating the necessity for Non-Maximum Suppression (NMS), thus diminishing inference latency and simplifying the object detection schema. The majority of object detection algorithms have been developed based on extensive image detection databases such as COCO and ImageNet, with training data encompassing over 100,000 images. The challenge lies in adapting these aforementioned architectures to specialized datasets, such as pavement damage, where the data available is comparatively limited.

The utilization of deep learning methodologies for the detection of road damage is primarily dependent on the RDD2022 database [12], an open-source repository. This comprehensive dataset contains 47,420 images of road damage, amassed from six distinct nations, and is annotated with over 55,000 instances of various damage types. Designed to bolster the CRDDC2022 challenge, the RDD2022 database is pivotal for the automated detection and classification of road damage via deep learning algorithms, thus serving as a crucial resource for road condition monitoring and the progression of computer vision research. Within the RDD2022 schema, road damage is methodically categorized into four main types: longitudinal cracks, which run parallel to the road's direction; transverse cracks, perpendicular to the road's course; alligator cracks, resembling an alligator's skin with interconnected fissures; and potholes, marked by depressions in the road surface. These classifications are integral to the training of deep learning models, facilitating the autonomous detection and assessment of road damage, which is essential for maintaining safer and well-maintained road infrastructures. Numerous academic publications have sought to refine deep learning approaches for pavement damage detection, with citations ranging from. These endeavors include efforts to augment the existing YOLO architecture by incorporating additional structures and applying diverse augmentation techniques to enhance its efficacy [13–20]. The classification of pavement damage detection was initially based on the original four categories of the RDD2022 challenge. However, practical applications necessitate more detailed road damage detection classes for the implementation of comprehensive road pavement damage analysis. Consequently, further research is imperative to expand the types of road damage and improve image detection capabilities. Nevertheless, practical pavement design requires an extended range of damage classifications that surpass the four types specified in the RDD2022 database.

In this study, we introduce a lightweight object detection model predicated upon the YOLOv9 architecture. The novel modification incorporates a partial attention block (PSA)[11] into the base model, which precedes the feature map tensor's progression to the YOLO detection head. Utilizing the RDD2022 database in conjunction with Thailand's pavement detection database, the model classifies seven unique types of pavement damage. The classifications include longitudinal wheel mark (D00), lateral crack (D10), alligator crack (D20), patching (D30), pothole (D40), crosswalk blur (D43), and white line blur (D44). A comprehensive comparison of the enhanced model's performance against the current state-of-the-art object detection models, namely YOLO8, YOLO9, and YOLO10, was conducted to verify its improved effectiveness. The principal contributions of this research are as follows:

- **Development of an Enhanced Object Detection Model:** This research introduces an advanced object detection model, which is predicated upon the architecture of the latest state-of-the-art models. It incorporates an attention mechanism into the model's architecture, thereby demonstrating augmented precision and expedited inference speeds surpassing those of previous models documented in the literature.

- **Comparative Evaluation with Specialized Dataset:** The study conducts a comprehensive evaluation of contemporary state-of-the-art object detection models, specifically YOLO8, YOLO9, and YOLO10, utilizing a specialized dataset for pavement damage. It is noteworthy that the size of this dataset is relatively limited when juxtaposed with the extensive COCO database.

- **Extension of Damage Detection Classification Range:** An expansion of the damage detection classification range has been realized, transcending the limitations of current state-of-the-art models. This extension offers a more veracious reflection of real-world scenarios pertinent to pavement engineering applications

## 2. Data characteristics

In the present study, pavement distress was systematically classified into seven discrete categories of damage. The primary category, designated as D00, pertains to longitudinal cracking, as illustrated in Figure 1. Longitudinal cracking, a prevalent form of pavement distress, is typified by fissures aligned parallel to the pavement's centerline or laydown direction. These cracks may stem from a variety of sources, including fatigue or top-down cracking, and their occurrence can precipitate moisture infiltration, thereby aggravating surface roughness and potentially heralding the onset of alligator cracking and structural failure, denoted as D20. The genesis of longitudinal cracks is complex, often tracing back to inadequate joint construction or improper placement. In contrast, lateral cracks in flexible pavements, identified as D10, are chiefly associated with fatigue cracking—a common mode of distress in such pavements. Fatigue cracking results from the interplay of repetitive, heavy traffic loads and thermal variations, which induce tensile stresses within the pavement matrix. Pothole formation in asphalt pavements is characterized by small, bowl-shaped indentations penetrating through the asphalt layer to the base course, with sharply defined edges and vertical walls at the depression's upper boundary. Potholes typically develop on roads with thin hot mix asphalt (HMA) overlays and are less frequent on routes with thicker HMA applications. When pavement damage occurs, the remedial strategy involves delineating the impacted zone and incorporating new asphaltic concrete through a procedure known as patching, coded as D30. This technique mitigates localized degradation by restoring the affected area with new material. Additionally, the detection of obscured or faded pavement markings, due to the associated traffic safety hazards, was of interest and is denoted as D44. Similarly, the imperceptibility of pedestrian crosswalks is indicated and coded as D43.

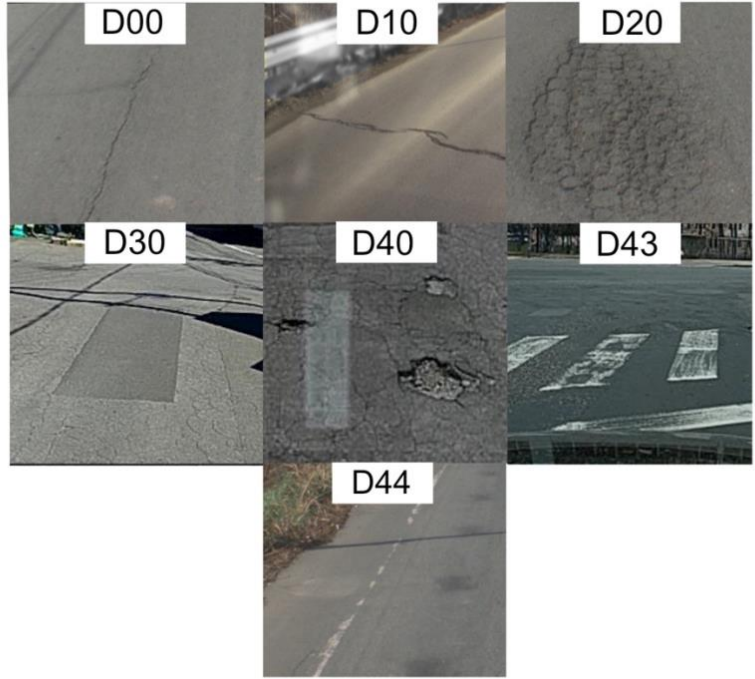

Figure.1 The road damage class

The dataset employed in this study was sourced from the RDD2022 database, enhanced by road detection data linked to the enterprise of the third author. It was further enriched by contributions from five countries, including an additional dataset from Thailand, depicted in Figure 2. The consolidated data was methodically categorized into seven class groups, as outlined in Table 1, following the guidelines set by the Highway Department of Thailand. This updated dataset version surpasses its predecessor by incorporating new categories relevant to current asphalt repair methods, notably 'patching', and clearly distinguishing pedestrian crossings and areas where white traffic lines are less visible. The inclusion of obscured or invisible crosswalks and lane markings adds a layer of complexity for the entity responsible for road upkeep. The data utilized to train the object detection deep learning model was not augmented to ensure an equitable comparison of the capacity of the proposed model with other state-of-the-art (SOTA) models.

The expansion of damage categories presents formidable challenges for deep learning frameworks, compelling the formulation of sophisticated classification algorithms to refine the prognostic accuracy of damage assessment. The dataset was partitioned into training, validation, and testing subsets to facilitate the model's training process (refer to Figure 3), following a distribution ratio of 80/10/10 (training/validation/testing). The validation subset played a pivotal role during the training phase for hyperparameter optimization, while the testing subset was instrumental in gauging the model's efficacy with novel data. A notable challenge encountered in the application of deep learning was the imbalance in the number of specified images per class. It was observed that the D40 and D43 categories exhibited a lower number of training instances, approximately 500 and 1000 respectively, which could potentially skew the learning process.

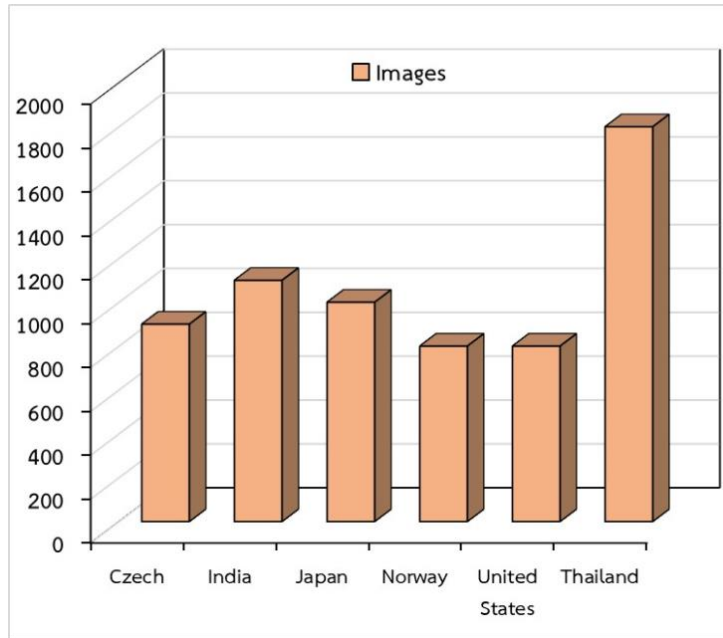

Figure 2 Source of image for training

Table 1 The class of road damage in this study

| Damage Type | | Detail | Class Name |
|---|---|---|---|
| Crack | Linear Crack | Longitudinal crack | D00 |
| | | Lateral crack | D10 |
| | Alligator Crack | Partial/Overall pavement | D20 |
| Patching | | Defective/Good Patching | D30 |
| Other Damage | | Pothole | D40 |
| | | Crosswalk blur | D43 |
| | | White line blur | D44 |

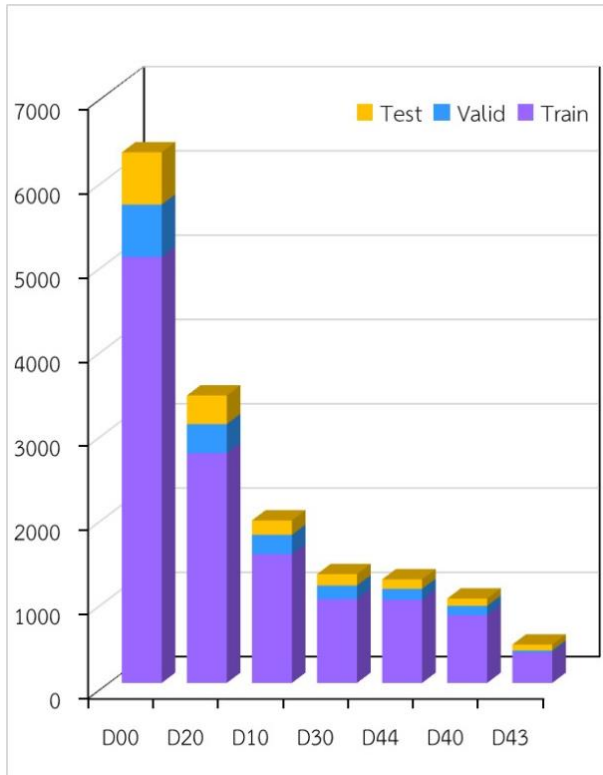

**Figure 3 Data label characteristics**

## 2. Model architecture

The model was developed based on the YOLO9s [10] architecture, which epitomizes the compact design within the YOLO9 series (see Fig. 4). Our objective was to engineer a lightweight model that preserves the efficiency of larger-scale models for real-time detection of road damage. The YOLO9s architecture is segmented into three primary components: the backbone, neck, and head. A distinctive feature of YOLO9s, compared to other larger models, is the absence of an auxiliary model part to assist in identifying the feature map during training. Nonetheless, critical components such as the Generalized Efficient Layer Aggregation Network (GELAN) are maintained within the model's architecture. An additional layer of partial attention was incorporated into the feature vector emanating from the neck segment of the model, prior to its transmission to the head segment for detection purposes. This concept of Partial Selective Attention (PSA) was adapted from the YOLO10 [11] architectural framework. We hypothesize that the PSA layer will enhance the identification of critical layers for road damage detection. This layer was applied to the deeper layers (#4 - #6) to facilitate the detection of edges or boundaries, which are the primary characteristics for identifying cracks on pavement, resulting in the variant named YOLO9tr.

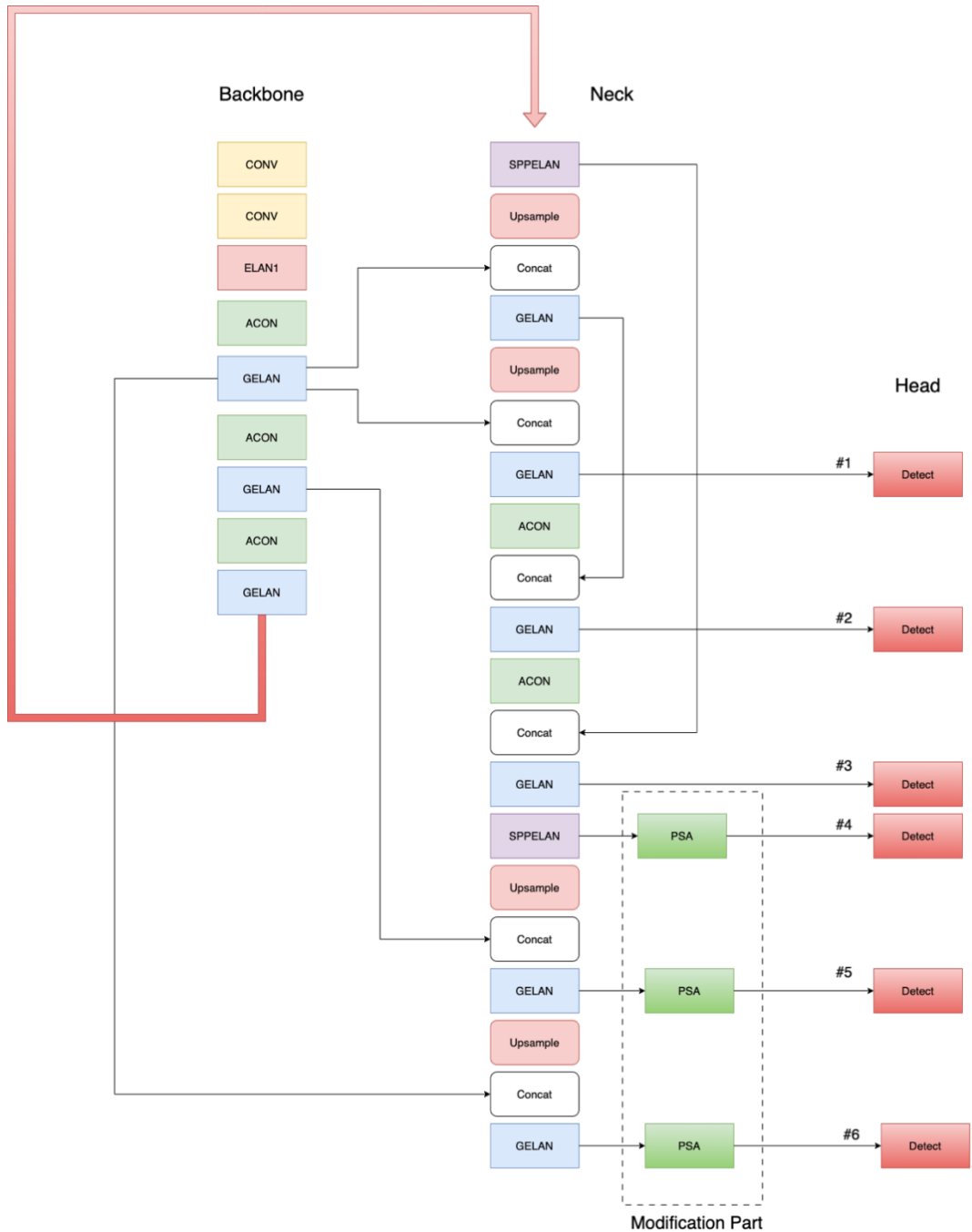

Figure 4 Model architecture improved model from YOLO9s

The architecture of partial attention block are shown in Fig. 5 and 6. The attention block (AB) architecture was the concept of the split the feature map into 3 components, query, key and value. Then it will be combined with the partial attention block (Fig. 6). The architecture of the attention block is defined by several key parameters that structure its functionality. The input tensor is $x \in \mathbb{R}^{B \times C \times H \times W}$. The attention block can be defined as follows:

$$Y = Conv(V \otimes Attention(Q,K,V).reshape(B,C,H,W) + Conv_{PE}(V.reshape(B,C,H,W))) \tag{1}$$

$$Q,K,V = Conv(x).view(B,M,D \cdot R \times 2 + D, N).split([D \cdot R, D \cdot R, D], dim = 2) \tag{2}$$

$$Attention(Q,K,V) = softmax(\frac{Q^T_K}{\sqrt{D \cdot R}})V \tag{3}$$

The batch size, denoted by B, is the number of samples processed in parallel during training or inference. The number of channels, C, indicates the depth of the feature map and is crucial for capturing the complexity of the input data. The height and width of the input feature map are represented by H and W, respectively, which together determine the spatial dimensions of the map. The total number of features within the map, N, is computed as the product of H and W. The dimension of each attention head, D, is a critical factor that influences the granularity of the attention mechanism. The attention ratio, R, is a parameter that adjusts the proportionality of the key dimensions in relation to the dimensions of the attention heads. Lastly, the number of attention heads, M, reflects the model's ability to simultaneously attend to various segments of the input feature map, enhancing its representational power. The output feature map ( Y ) is obtained by reshaping and projecting the attended features back to the original dimensions: Our model employs several key operations: *Conv()* and *Conv_{PE}()* for convolution, *softmax()* for normalizing attention scores, and *split()* to separate tensor components.

The architecture of the partial attention block is delineated in Figure 6. This architecture integrates a convolutional framework with batch normalization and an Attention Block (AB). Initially, the feature map traverses the AB, where it is bifurcated into two segments. The segment on the right is subsequently channeled into the AB and a convolutional layer. A residual connection is established between the input of the block and the feature map transiting through the block, amalgamated to mitigate the attenuation of gradients that typically occurs in extensive deep learning architectures.

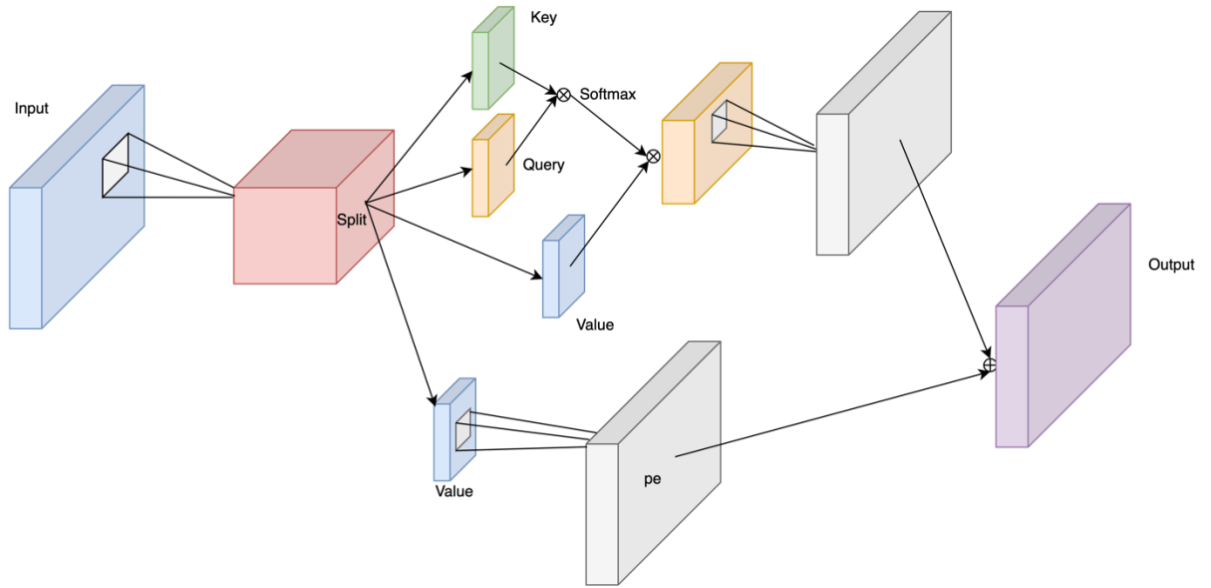

Figure 5 The attention block (AB)

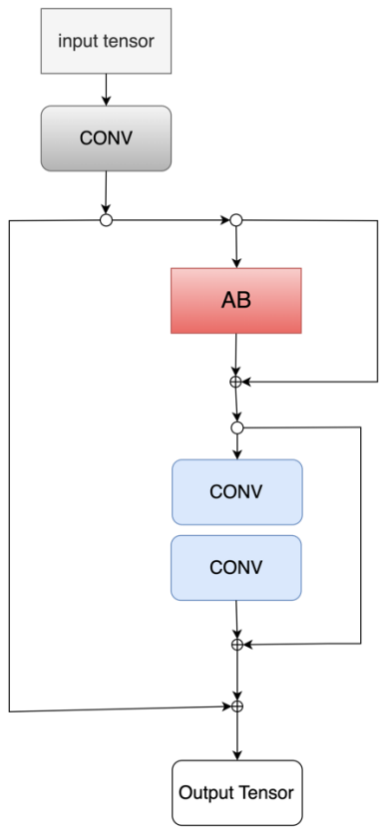

Figure 6 The partial attention block (PSA)

## 4. Experiment

The study employed the Vast Ai Cloud computational platform [21], equipped with dual NVIDIA GeForce RTX 4090 GPUs. The deep learning architecture's construction was enabled through the PyTorch framework (version 2.2.0), while CUDA version 12.1 was utilized as the programming language for the compilation of the code. The YOLOv9 object detection algorithm was sourced from a GitHub repository [22]. Throughout the training phase of the model, mosaic data augmentation was applied at an intensity level of 1.0, and the mixup algorithm was incorporated with a parameter value of 0.5. Input images were standardized to a resolution of 640×640 pixels. The batch size was adjusted based on the maximum memory capacity available. A total of 200 epochs were allocated for training, with a focus on optimizing the model's validation performance. The model's weights were saved at the epoch which yielded the best validation results. The hyperparameter that controls the model's attention ratio, denoted as R, was set at 0.5. The number of attention heads was determined by the input channel's dimensions.

### 4.1 Indicators of evaluation

In the conducted research, the evaluation of the recognition efficacy for road surface damage was performed using a set of metrics: precision, recall, F1-score, FPS (frames per second), and mean average precision (mAp). Precision (P) quantifies the classifier's accuracy in predicting road surface damage, denoted by the ratio of true positive instances to the overall positive instances identified by the classifier. Conversely, recall (R) measures the classifier's ability to detect all instances of road surface damage, represented by the ratio of accurately identified positive instances to the total actual positive instances. The mathematical expressions for precision and recall are delineated in Equations (4) and (5), respectively:

$$P = \frac{TF}{TF + FP} \qquad (4)$$

$$R = \frac{TF}{TF + FN} \qquad (5)$$

In this context, *TP* represents the count of accurately identified positive instances (true positives), *FP* indicates the quantity of falsely identified positive instances (false positives), FN refers to the number of negative instances incorrectly classified (false negatives), and TN stands for the count of negative instances correctly classified (true negatives).

The mean average precision (mAp) serves as an indicator of the detection precision within target recognition tasks. The mAp is derived by computing the mean of the precision values for each category, which is obtained through the integration of the Precision–Recall (P–R) curve, and subsequently averaging these values. The equation for *mAP* is presented in Equation (6):

$$\text{mAp} = \frac{1}{N}\sum_{i=1}^{N} \text{AP}_i \qquad (6)$$

The F1-score is a harmonizing metric that equilibrates the measures of precision and recall. This metric is designated as the evaluative standard in the IEEE Big Data 2022 [48] Road Damage Detection Challenge. It is mathematically articulated as follows in Equation (7):

$$F1\text{-}score = 2 \times \frac{P \times R}{P + R} \qquad (7)$$

The speed of detection was very important for the object detection of deep learning. The frame per second (FPS) was use as one of the indicator index in this study

$$FPS = \frac{1}{t^{pre} + t^{inference} + t^{post}} \qquad (8)$$

The $t^{pre}$ represents the time allocated for preprocessing prior to detection. $t^{inference}$ denotes the duration of the inference process. $t^{post}$ is the time subsequent to inference.

### 4.2 Experiment Results

The results of the model's training are detailed in Table 2. The precision metrics for the model displayed minimal variation among the different classes. Notably, class D43, which corresponds to blurred crosswalks, registered the highest precision. In contrast, the precision for detecting blurred white lines was the lowest. The detection of potholes and patching, classes D40 and D30, respectively, showed high precision. These were more discernible and straightforward to detect compared to other types of pavement damage, such as cracks. Particularly challenging was distinguishing between alligator cracks and a series of longitudinal cracks. The recall for alligator cracks, class D20, was the lowest among the types of failures detected. Precision for other categories of pavement damage was relatively consistent. The mean Average Precision (mAp50) of our study was significantly lower than that reported in previous research. This difference may be attributed to the greater diversity of detection classes and the unbiased, random selection of images used in our study. In some cases, the detection was so challenging that it bordered on the limits of human visual judgment.

Table 2 The test results of proposed model (YOLO9tr)

|         | mAp50 | P     | R     |
|---------|-------|-------|-------|
| Overall | 0.655 | 0.652 | 0.616 |
| D00     | 0.688 | 0.591 | 0.643 |
| D10     | 0.575 | 0.614 | 0.573 |
| D20 | 0.579 | 0.630 | 0.466 |
| D30 | 0,708 | 0.656 | 0.644 |
| D40 | 0.719 | 0.715 | 0.667 |
| D43 | 0.774 | 0.718 | 0.746 |
| D44 | 0.543 | 0.537 | 0.575 |

In Table 3, Figs 7 and 8, we present a comparative analysis of our proposed model against current state-of-the-art (SOTA) models. To maintain the integrity of the comparison, neither our proposed model nor the YOLOv9 series underwent reparameterization. The proposed models, YOLO9tr and YOLO9tr-L, exhibited enhanced performance in terms of mean Average Precision (mAp50), achieving an equilibrium between model dimensions and inference velocity. Our models paralleled the mAp50 values of larger models, such as RT-Deter, YOLO8x, and YOLO9e, which possess extensive parameter sets. However, these models necessitate substantial computational resources and extended inference time, resulting in diminished frame rates for image detection. In contrast, our models are comparable in size to smaller SOTA models like YOLO10s and YOLO8s, yet they significantly outperform them in precision. A salient feature of our proposed models is their minimal parameterization, facilitating a high frame rate of up to 136 frames per second (FPS) in image detection tasks. This capability is especially advantageous for video applications where high frame rates are not requisite.

Table 3 Comparison to the previous model

| Model | mAp$_{50}$ | F1-score | Parameters (Million) | FLOPs G | FPS |
|---|---|---|---|---|---|
| RT-Deter | 0.654 | 0.640 | 76 | 259 | 74 |
| YOLO8s | 0.592 | 0.600 | 11.2 | 28.6 | 313 |
| YOLO8m | 0.588 | 0.590 | 25.9 | 78.9 | 227 |
| YOLO8l | 0.602 | 0.590 | 43.7 | 165.2 | 152 |
| YOLO8x | 0.590 | 0.53 | 68.2 | 257.8 | 109 |
| YOLOv10s | 0.533 | 0.560 | 8.0 | 24.5 | 454 |
| YOLOv10b | 0.570 | 0.590 | 20.4 | 98.0 | 166 |
| YOLOv10x | 0.557 | 0.570 | 31.6 | 169.8 | 117 |
| YOLO9m | 0.649 | 0.630 | 32.6 | 130.7 | 128 |
| YOLO9c | 0.636 | 0.620 | 50.7 | 236.7 | 108 |
| YOLO9e | 0.658 | 0.640 | 68.6 | 240.7 | 75 |
| YOLO9etr | 0.653 | 0.630 | 75.8 | 250 | 73 |
| YOLO9ctr | 0.651 | 0.630 | 69.3 | 228 | 75 |
| **YOLO9tr-L** | **0.658** | **0.640** | **10.6** | **42.6** | **117** |
| YOLO9tr | 0.655 | 0.630 | 10.2 | 41.1 | 136 |

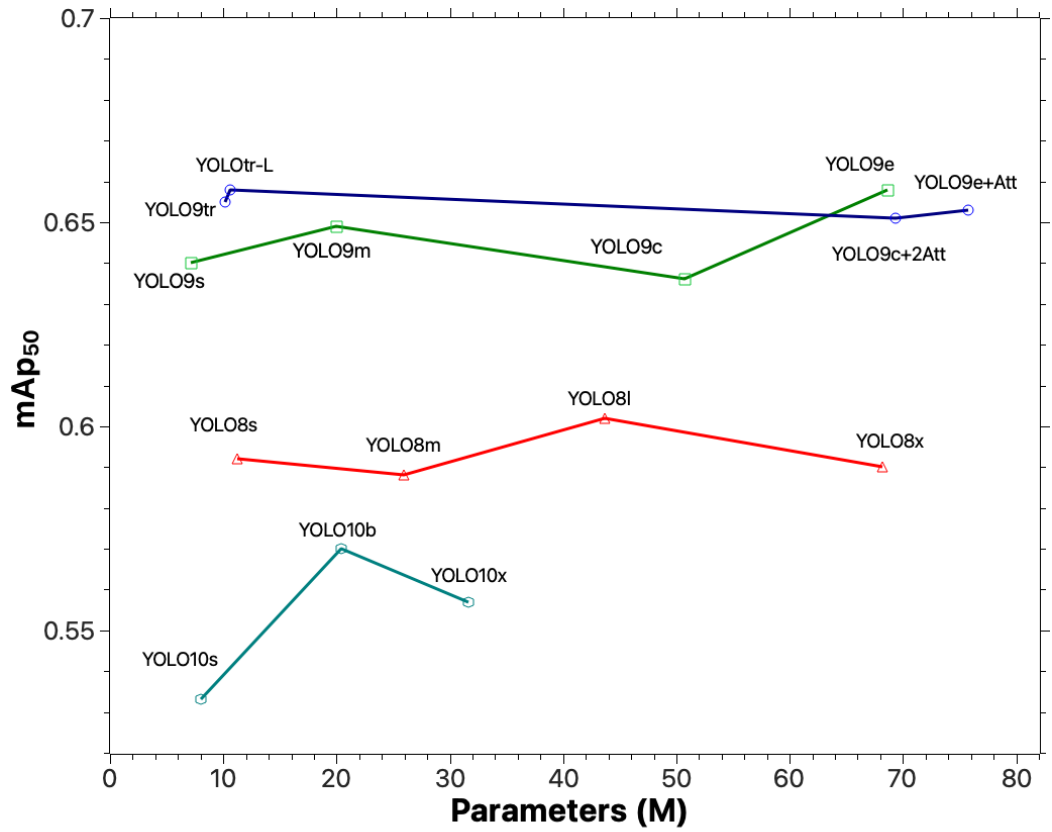

**Figure 7** The comparison between mAp$_{50}$ of different model with different parameters

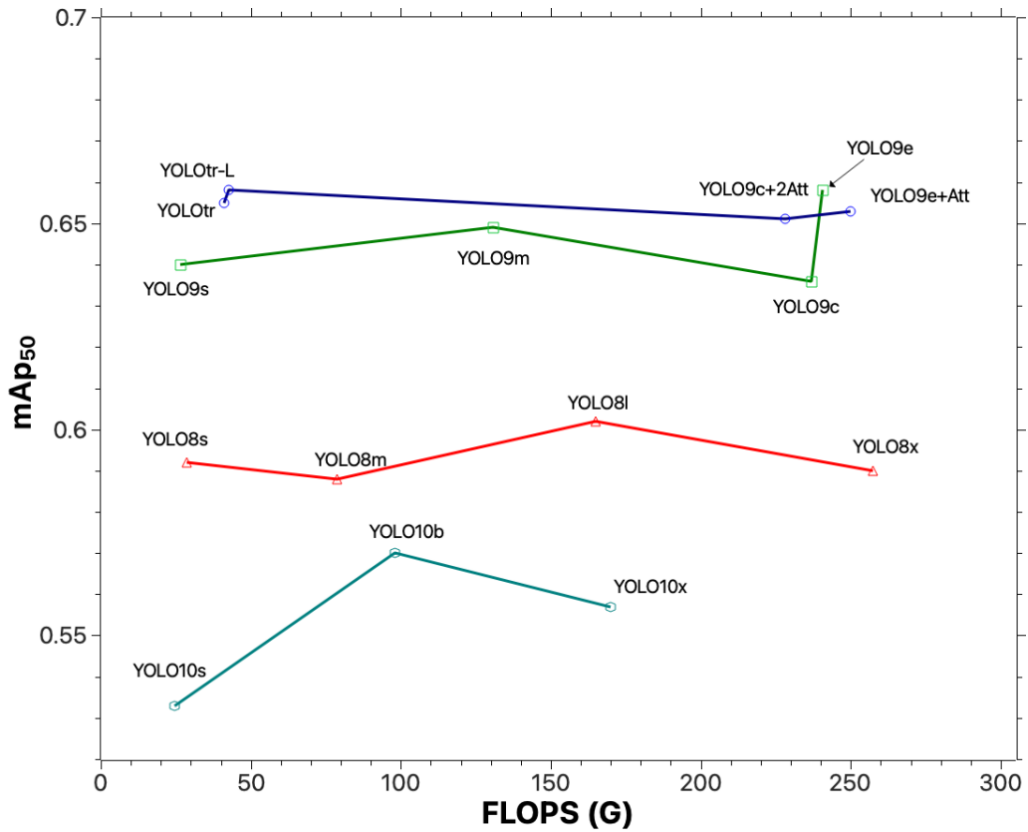

**Figure 8** The comparison between mAp50 of different model with FLOPS (G)

## 4. Visualization

The comparative analysis of image detection capabilities between the proposed model and the state-of-the-art (SOTA) model is illustrated in Figure 9. The images, sourced from various countries, demonstrate the models' performance under diverse circumstances and camera angles for pavement damage detection. It is evident that our proposed model discerns finer details compared to other competitive models. Specifically, our model successfully identifies the D00 damage category in the leftmost image, a feat that other models, such as YOLO9e and YOLO8x, fail to achieve. Moreover, in the challenging scenario of a blurred crosswalk depicted in the second image, our proposed model exhibits robust detection capability.

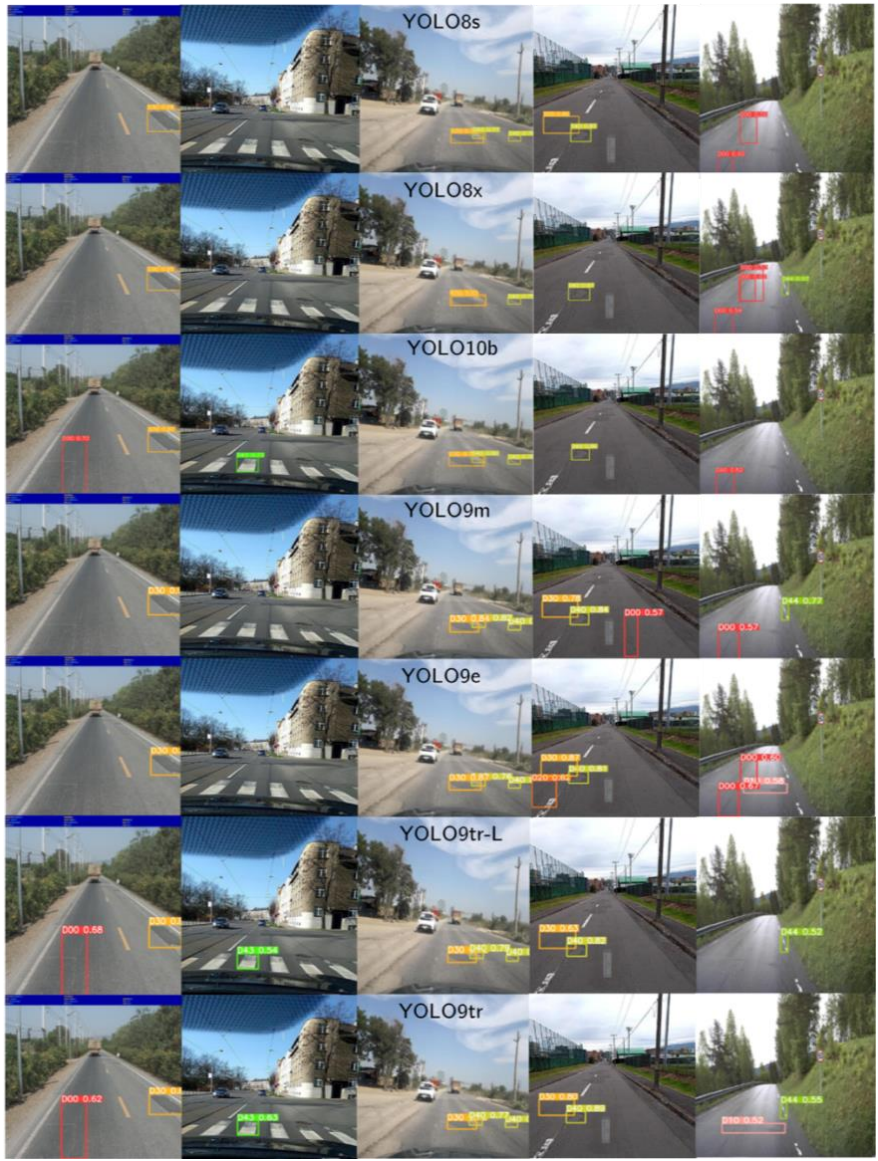

Figure 9 Detection results of different models

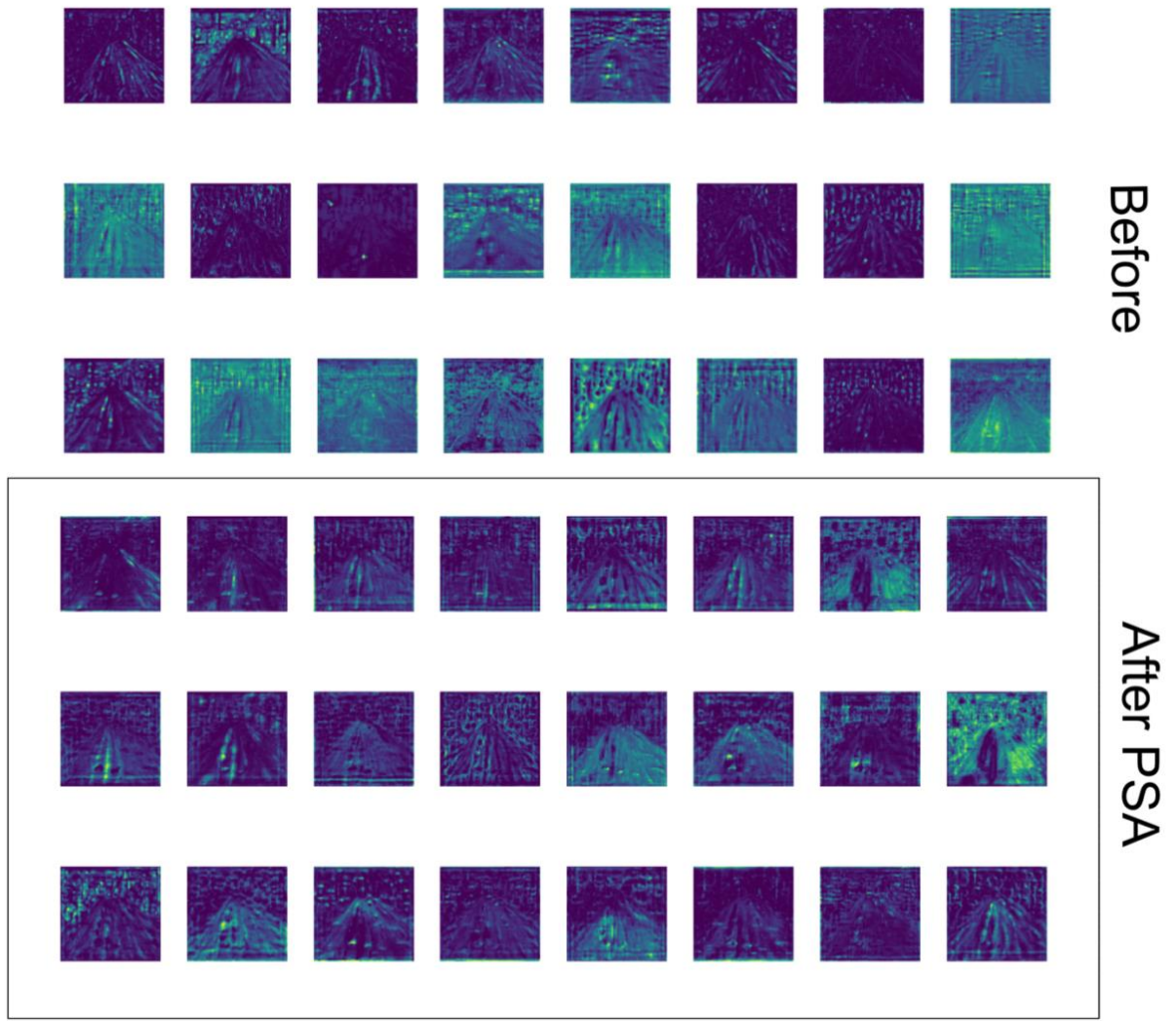

Figure 10 The feature map of model before and after applied partial self-attention layer in model

The feature map tensor before and after applied PSA layer (#6) is shown in Fig. 10. The application of a Partial Self-Attention (PSA) layer significantly enhances the quality of feature maps in a neural network, as evidenced by the comparison of feature maps before and after the PSA layer is applied. Initially, the feature maps are characterized by distinct, noisy, and scattered spatial distributions, indicating the capture of redundant or irrelevant information and less correlation or interaction between features. Post-PSA, the feature maps exhibit more uniformity and smoothness, with refined patterns and reduced noise, suggesting that the PSA layer effectively enhances the consistency and coherence of feature representations. This transformation implies a better integration and correlation of features, leading to more holistic and robust representations. The PSA layer's ability to focus on more relevant regions reduces noise and emphasizes important features, which is crucial for tasks such as object detection or segmentation. This improvement aligns with findings in the literature where self-attention mechanisms have been shown to enhance feature learning by enabling networks to weigh the importance of different spatial locations more effectively. Consequently, the network can capture and utilize critical features more efficiently, potentially leading to improved overall performance.

## 5. Ablation Study

The current investigation undertook an ablation study to delineate the ramifications of architectural alterations and hyperparameter adjustments on the model's performance. Employing YOLO9s as the baseline, it provided a benchmark against which the augmented model, integrated with PSA layers, was evaluated. The experimental framework encompassed a comparative analysis between a single attention layer, illustrated at position #6 in Figure 3, and a complex attention architecture consisting of six layers. This methodology required each feature map to be processed through the PSA before advancing to the YOLO detection head. Furthermore, the attention ratio parameter was escalated from its initial setting of 0.5 in the prototype to 1.0, aiming to appraise its influence on the model's functional efficiency. Throughout the comprehensive ablation study, the proposed model manifested superior precision and inference velocity. The YOLO9tr-L model exhibited a marginally elevated mAp50 score. Nonetheless, its inference speed and training duration surpassed that of the proposed model. The YOLO9s, augmented with a solitary attention step, significantly underperformed in mAp50 compared to the unmodified YOLO9s. Efforts to modify larger models, namely YOLOe and YOLOc, by incorporating the PSA layer, did not markedly enhance performance. In fact, the YOLOe modification yielded slightly inferior performance relative to the original model.

Table 4 the results for the abrasion study

| Models | mAp50 | F1-score | FLOPS(G) |
|---|---|---|---|
| YOLO9s | 0.640 | 0.620 | 38.7 |
| YOLO9tr-Single Attention layer | 0.587 | 0.570 | 40.1 |
| YOLO9tr-L-6 Attention layer | 0.658 | 0.640 | 42.6 |
| YOLO9etr-1 Attention layer | 0.653 | 0.630 | 236.7 |
| YOLO9ctr- -2 Attention layer | 0.651 | 0.630 | 240.7 |
| YOLO9tr-R=1.0 | 0.647 | 0.630 | 41.2 |
| YOLO9tr | **0.655** | **0.630** | **41.1** |

## 6. Discussion

The results from the YOLOv9tr model underscore its capability to efficiently and accurately detect pavement damage, advancing the state-of-the-art in object detection. By incorporating a partial attention block into the YOLOv9 architecture, the model leverages enhanced feature extraction and attention mechanisms, leading to improved detection performance, particularly in complex scenarios such as blurred images or intricate damage patterns. In a comparative analysis with contemporary models like YOLO8 and YOLO9, YOLOv9tr consistently demonstrated superior precision and inference speed. Specifically, the YOLOv9tr achieved a balance between computational efficiency and detection accuracy, a crucial factor for real-time applications. The model's precision, particularly in identifying the D00 damage category and maintaining robustness in blurred conditions, outperformed models like YOLO9e and YOLO8x, which struggled with these challenging scenarios. The YOLOv9tr model's capability to process up to 136 frames per second (FPS) positions it as an ideal candidate for real-time monitoring systems. This high frame rate is particularly beneficial for applications in video surveillance and automated inspection systems, where timely detection and response are critical. Additionally, the model's compact architecture, with minimal parameterization and computational load, ensures its deployment feasibility on devices with limited processing power.

An essential contribution of this research is the expansion of the damage classification range, addressing the limitations of existing models that typically categorize only four types of damage. By extending the classifications to include seven types, such as longitudinal wheel marks, lateral cracks, and crosswalk blurs, YOLOv9tr provides a more comprehensive and realistic assessment of pavement conditions. This expanded classification capability is pivotal for pavement engineering applications, where diverse damage types need to be accurately identified and assessed for maintenance and safety purposes. Future research can further optimize the YOLOv9tr model by exploring additional augmentation techniques and integrating more sophisticated attention mechanisms. Moreover, applying this model to other domains requiring real-time object detection, such as autonomous driving or security monitoring, could validate its versatility and robustness. Additionally, expanding the dataset to include more diverse damage scenarios from various geographical regions could enhance the model's generalization capabilities, ensuring its effectiveness in a broader range of real-world applications. In summary, the YOLOv9tr model represents a significant step forward in object detection technology, particularly for applications in pavement damage detection. Its balanced performance, real-time processing capability, and expanded classification range highlight its potential for practical deployment and set a new benchmark for future research in this domain.

## 7. Conclusion

In conclusion, the YOLOv9tr model demonstrates significant advancements in object detection, particularly in pavement damage detection tasks. Our model exhibits a balanced trade-off between precision, model size, and inference speed, making it suitable for real-time applications. The YOLOv9tr model's high frame rate of up to 136 FPS is particularly advantageous for video applications, outperforming larger models like RT-Deter, YOLO8x, and YOLO9e in terms of precision while maintaining a compact size akin to smaller models such as YOLO10s and YOLO8s.

The comparative analysis underscores the superior performance of YOLOv9tr in detecting finer details and handling challenging scenarios, such as blurred images, more effectively than other state-of-the-art models. Our findings suggest that the YOLOv9tr model holds substantial potential for practical deployment in real-time pavement condition monitoring and other similar applications. The success of the YOLOv9tr model in maintaining high precision with minimal parameterization and computational resource requirements underscores its efficacy and utility. Future work could explore further optimizations and the application of this model to other domains requiring efficient and accurate real-time object detection